\documentclass{article}

\usepackage{microtype}
\usepackage{graphicx}
\usepackage{subcaption}
\usepackage{booktabs}
\usepackage{multirow}
\usepackage{makecell}

\usepackage{hyperref}

\usepackage[preprint]{icml2026}
\usepackage{amsmath}
\usepackage{amssymb}
\usepackage{mathtools}
\usepackage{amsthm}
\usepackage{enumerate}
\usepackage{enumitem}
\usepackage[capitalize,noabbrev]{cleveref}

\theoremstyle{plain}

\theoremstyle{definition}

\theoremstyle{remark}

\usepackage{listings}
\usepackage{xcolor}
\usepackage[textsize=tiny]{todonotes}
\usepackage{tcolorbox}
\usepackage{seqsplit}

\icmltitlerunning{Fragment-Oriented Ranking and Generation}

\begin{document}
\newcommand{\icmlCorrespondingAuthor}{\textsuperscript{\textdagger}Corresponding author.}
\twocolumn[
  \icmltitle{FORGE: Fragment-Oriented Ranking and Generation for Context-Aware Molecular Optimization}

  \icmlsetsymbol{equal}{*}
  \icmlsetsymbol{cor}{\textdagger}
  
  \begin{icmlauthorlist}
    \icmlauthor{Qingchuan Zhang}{ustc}
    \icmlauthor{He Cao}{idea}
    \icmlauthor{Hao Li}{pku}
    \icmlauthor{Yanjun Shao}{yale}
    \icmlauthor{Zhiyuan Liu}{nus}
    \icmlauthor{Shihang Wang}{mpu}
    \icmlauthor{Shufang Xie}{zgc}
    \icmlauthor{Shenghua Gao}{hku}
    \icmlauthor{Xinwu Ye}{hku,cor}
\end{icmlauthorlist}
    
    \icmlaffiliation{ustc}{University of Science and Technology of China}
    \icmlaffiliation{idea}{International Digital Economy Academy}
    \icmlaffiliation{pku}{Peking University}
    \icmlaffiliation{yale}{Yale University}
    \icmlaffiliation{nus}{National University of Singapore}
    \icmlaffiliation{mpu}{Macao Polytechnic University}
    \icmlaffiliation{zgc}{Zhongguancun Academy}
    \icmlaffiliation{hku}{University of Hong Kong}
    
    \icmlcorrespondingauthor{Qingchuan Zhang}{misaka@mail.ustc.edu.cn}
    \icmlcorrespondingauthor{Xinwu Ye}{xinwuye43@connect.hku.hk}
  \icmlkeywords{FORGE: Context-Aware Fragment Ranking and Modification for Molecular Optimization with a 0.6B Language Model}
]

\printAffiliationsAndNotice{\icmlCorrespondingAuthor}

\begin{abstract}
Molecular optimization seeks to improve a molecule through small structural edits while preserving similarity to the starting compound. Recent language-model approaches typically treat this task as prompt-conditioned sequence generation. However, relying on natural language introduces an inherent data-scaling bottleneck, often leads to chemical hallucinations, and ignores the strong context dependence of fragment effects. We present FORGE, a two-stage framework that reformulates molecular optimization as context-aware local editing. By utilizing automatically mined, verified low-to-high edit pairs instead of expensive human text annotations, Stage 1 ranks candidate fragments by their property contribution under the full molecular context to inject chemical prior, and Stage 2 generates explicit fragment replacements. Built on a compact 0.6B language model, FORGE further adapts to unseen black-box objectives through in-context demonstrations. Across Prompt-MolOpt, PMO-1k and ChemCoTBench, FORGE consistently outperforms prior methods, including substantially larger language models and graph methods. These results highlight the value of explicit fragment-level supervision as a more easily obtainable, scalable, and hallucination-less alternative to natural language training.
\end{abstract}

\section{Introduction}
Molecular optimization aims to improve a starting molecule toward a desired property, often under a strict structural similarity constraint~\citep{jin2018junction, you2018graph, dara2022machine}. Unlike open-ended molecular generation, molecular optimization is fundamentally a \emph{local editing} problem: the goal is not to sample an arbitrarily high-scoring molecule, but to improve a lead compound through small structural modifications~\citep{jin2018learning, jin2019hierarchical}. This local-edit regime underlies many practical drug-discovery workflows~\citep{zhang2024deep, ferreira2025ai} as well as standard benchmarks such as similarity-constrained optimization and black-box oracle-guided search~\citep{gao2022sample, brown2019guacamol}.

Chemistry language models have made SMILES generation a common interface
for molecular design \citep{chithrananda2020chemberta, guevorguian2024small,
yu2024llasmol}, and most recent optimization methods follow the same
pattern: given an input molecule and a textual description, decode
an improved molecule. We argue that this formulation is poorly matched to
the local-edit setting, in two specific ways: \textbf{(i) Semantic mismatch and weak prompt control}: Real-world drug discovery oracles (e.g., ADMET classifiers, proprietary binding scores) lack faithful natural-language descriptions, resulting in weak prompt control. Training on LLM-distilled chain-of-thought data may partially fill the gap, but tends to
inherit known chemical inaccuracies from the teacher model. \textbf{(ii) Ignored context dependence}: Fragment contributions depend strongly on their local environment. Representing substructure attribution as a global, context-free scalar collapses critical structural information and introduces label noise that propagates to training.

We therefore reformulate molecular optimization as \emph{context-aware local editing}: Rather than asking a model to rewrite the whole molecule from a property prompt, we decompose optimization into two explicit decisions: \emph{where} to edit and \emph{how} to edit it. We instantiate this idea in \textbf{F}ragment-\textbf{O}riented \textbf{R}anking and \textbf{GE}neration (\textbf{FORGE}), a two-stage framework built on a compact 0.6B language model (Qwen3-0.6B~\citep{yang2025qwen3}) equipped with atom-level tokenization for stable fragment identity. During training, FORGE relies on rule-based data rather than LLM distillation. Stage~1 localizes edits by ranking fragments based on their context-conditioned contribution, and Stage~2 generates explicit fragment replacements. Stage~2 supervision combines rule-based and predictor-based attribution signals with \textbf{SME+}, our context-conditioned extension of \textbf{S}ubstructure \textbf{M}ask \textbf{E}xplanation (SME)~\citep{wu2023chemistry} for fragment attribution and pair construction. At inference, FORGE adapts to unknown black-box oracles through in-context demonstrations. 

Two empirical observations support this design and are detailed in Section~\ref{sec:motivation}: replacing the true target property in a PMO
prompt with an unrelated or unknown name changes the optimization score by
less than 3\%, and conditioning fragment attributions on local context
reduces attribution variance by 8\%--10\% across two
independent predictors. 
Our contributions are as follows:
\vspace{-0.4cm}
\begin{itemize}[leftmargin=10pt, noitemsep]
    \item We show that context-conditioned, rule-verified fragment edits
      are viable and highly scalable training data for molecular optimization, removing the
      need for rare natural-language objectives or expensive LLM-distilled chain-of-thought
      data.
    \item We propose \textbf{FORGE}, a two-stage framework that decomposes optimization into \emph{where} to edit and \emph{how} to edit, using context-conditioned fragment ranking and explicit fragment replacement generation, together with atom-level tokenization for stable fragment identity.
    \item We show that FORGE achieves state-of-the-art results across similarity-constrained, black-box, and real-target molecular optimization benchmarks using only a 0.6B model, outperforming larger baselines.
\end{itemize}

\section{Related Work}
\label{sec:related}

\paragraph{Fragment-based optimization and substructure attribution.}
Matched molecular pair analysis (MMPA)~\citep{doi:10.1021/acs.jcim.8b00173} treats local edits as the basic unit of medicinal chemistry, but its core assumption that fragment effects are only weakly context-dependent breaks down at activity cliffs~\citep{van2022exposing}. Deep-learning variants such as JTNN~\citep{jin2018junction}, GCPN~\citep{you2018graph}, Modof~\citep{chen2021deep}, Graph Polish~\citep{ji2021graph}, Prompt-MolOpt~\citep{wu2024leveraging}, and fragment-library methods including fRAG~\citep{lee2024molecule} and GenMol~\citep{lee2025genmol} still hard-code fragment edits as graph operators or external vocabularies, typically averaging fragment statistics across contexts. In parallel, substructure attribution methods such as SME~\citep{wu2023chemistry}, fragment-level Shapley values~\citep{doi:10.26434/chemrxiv.15002302/v1}, B-XAIC~\citep{proszewska2025b}, and fragment-wise attribution~\citep{musial2025fragment} improve the reliability of fragment importance estimates. Our method differs from these lines of work: we promote fragments to first-class editing units and make their contribution explicitly context-conditional; correspondingly, SME+ keeps the attribution mechanism of SME but changes the downstream aggregation scheme from global min--max scaling to within-context binning.


\paragraph{Chemistry language models and inference-time adaptation.}
SMILES-based pretraining and instruction tuning have established language modeling as a competitive route for molecular modeling, from ChemBERTa~\citep{chithrananda2020chemberta} and GP-MoLFormer~\citep{ross2025gp} to ChemLactica~\citep{guevorguian2024small}, ChemFM~\citep{cai2025chemfm}, and LlaSMol~\citep{yu2024llasmol}; related efforts revisit pretraining protocols~\citep{chitsaz2025novomolgen} or latent-space reasoning~\citep{ye2026latentchem}. Our approach departs from this line in two ways: we use atom-level tokenization rather than BPE (\cref{sec:qwenatom}), and we explicitly decompose optimization into discrimination and generation, allowing a 0.6B model to match or outperform 8B reasoning baselines and GPT-4o~\citep{hurst2024gpt}. More broadly, our inference procedure is motivated by recent progress in in-context learning (ICL) for chemistry, where methods like MOLLEO\citep{2024arXiv240616976W} and DemoDiff~\citep{liu2025graph} show strong gradient-free adaptation. Building on these insights, our method relies on dynamic in-context learning throughout the search process to adapt to unseen oracles. FORGE conditions on the target objective only through demonstrations sampled at search time, avoiding costing gradient-based test-time updates.
\section{Motivation Analysis}
\label{sec:motivation}

\begin{figure}[!ht]
\centering
\begin{subfigure}[b]{0.49\columnwidth}
  \includegraphics[width=\linewidth]{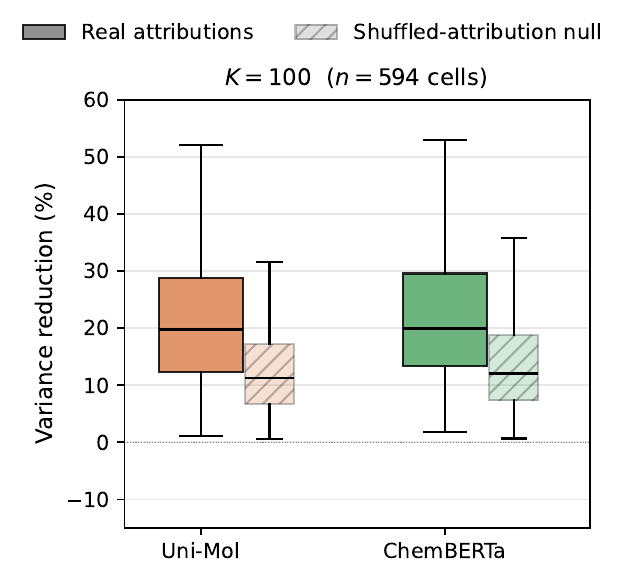}
  \caption{Context grouping reduces attribution variance.}
  \label{fig:motivation-a}
\end{subfigure}\hfill
\begin{subfigure}[b]{0.49\columnwidth}
  \includegraphics[width=\linewidth]{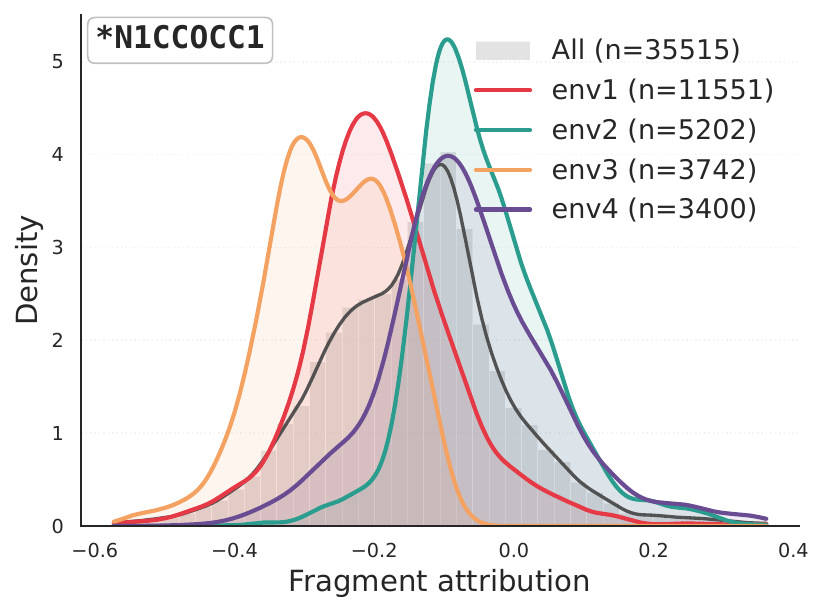}
  \caption{Context-dependent fragment effects of ESOL.}
  \label{fig:motivation-b}
\end{subfigure}\\[3pt]
\begin{subfigure}[b]{0.9\columnwidth}
  \includegraphics[width=\linewidth]{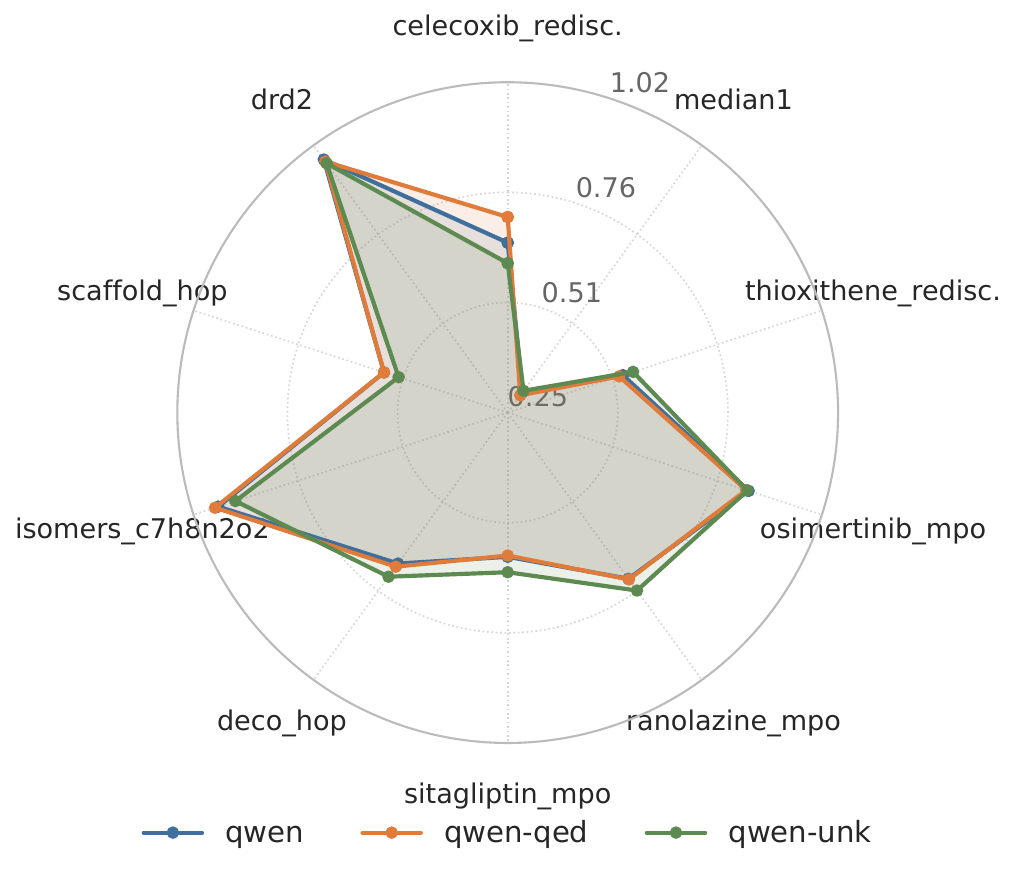}
  \caption{Weak prompt grounding on PMO.}
  \label{fig:motivation-c}
\end{subfigure}
\vspace{-0.1cm}
\caption{\textbf{Why prompt-only generation is insufficient for molecular optimization.}
\textbf{(a)} Grouping fragment attributions by ECFP neighborhoods reduces attribution variance across two distinct property predictors, indicating that fragment effects depend on chemical context.
\textbf{(b)} The same fragment can have different effects across host molecules.
\textbf{(c)} On PMO, replacing the true target property in the prompt with an unrelated or unknown one barely changes the optimization behavior of Qwen3-30B-A3B, indicating weak prompt grounding of molecular objectives.}
\vspace{-0.5cm}
\label{fig:motivation}
\end{figure}

We motivate FORGE with two observations. First, in recent LLM-based optimization pipelines, the target objective might only be weakly determined by its natural-language description. Second, fragment effects are strongly context-dependent, making context-free fragment statistics a poor supervision signal for local editing. Figure~\ref{fig:motivation} summarizes both effects. 

\subsection{Prompts weakly specify black-box objectives}
If prompt semantics reliably control molecular optimization, then changing the property description should alter optimization behavior. We test this on the 22 PMO tasks by running Qwen3-30B-A3B \citep{yang2025qwen3} inside the MOLLEO search loop \citep{2024arXiv240616976W} under three prompts: the default task-specific prompt, a mismatched prompt that always asks to optimize `QED', and a prompt that asks to optimize an `unknown property'. We keep the model, search loop, decoding, and oracle fixed, changing only the prompt description.
The result is stable: replacing the true objective with an unrelated or unknown description changes the aggregate PMO score by less than $3\%$, and on 13 of 22 tasks the per-task difference stays within $\pm 0.05$. As shown in Figure~\ref{fig:motivation-c}, prompt text alone is therefore a weak control signal for PMO-style black-box optimization. In this setting, useful task information should instead come primarily from oracle feedback and demonstrations.

\subsection{Fragment effects depend on molecular context}
\label{sec:vr}
Prompt text alone does not tell us what the model should condition on. A natural alternative is the fragment identity itself, but we find that fragment utility depends strongly on its molecular context: the same substructure can help in one molecule and hurt in another.

We test this through fragment attribution. Let $f:\mathcal{X}\rightarrow\mathbb{R}$ be a scalar property predictor or oracle, and let a molecule $x$ be decomposed into fragments $\{g_1,\dots,g_K\}$. For a fragment $g$ in host molecule $x$, we define its attribution as
$
a(g;x)=f(x)-f(x\setminus g),
$
where $x\setminus g$ is obtained either by masking $g$ (for GNN predictors) or by explicit deletion and recomputation (for rule-based properties). This quantity is a local perturbation proxy rather than a causal effect estimate, but it is enough to ask whether fragment utility varies systematically across host contexts. To quantify the contribution of local context, we group the occurrences of each fragment by its local Extended-Connectivity Fingerprints (ECFP) environment $e_r(g;x)$ at the attachment atoms and measure the resulting variance reduction,
\begin{equation}
\mathrm{VR}(g)=1-\frac{\sigma_{\mathrm{grouped}}}{\sigma_{\mathrm{original}}}.
\end{equation}
A positive $\mathrm{VR}(g)$ means that conditioning on local context reduces the spread of the fragment's observed effects.

Table~\ref{tab:vr-main} reports the VR result for Uni-Mol \citep{zhou2023unimol} and ChemBERTa \citep{chithrananda2020chemberta} trained on MoleculeNet\citep{2017arXiv170300564W}, with groups shuffled randomly as contrast. The pattern is consistent across settings: context grouping always reduces attribution variance, and the gap against the shuffled grouping remains $8\%$--$11\%$. Figure~\ref{fig:motivation}(a,b) shows the same effect.


The choice of neighborhood radius controls how finely context is partitioned. Table~\ref{tab:vr-radius} sweeps $r{=}1$ to $r{=}4$ on ChemBERTa: the genuine signal $\Delta$ grows monotonically with $r$, but larger radii also fragment the data into many more environment cells (avg.\ envs grows from 7.5 to 493.5), leaving too few samples per cell for downstream pair construction. We use $r{=}2$ throughout the rest of the paper as a practical trade-off between decoupling strength and per-cell sample size; per-property breakdowns are in Appendix~\ref{app:vr}. Extrapolating this monotonic trend leads to a natural theoretical limit: $r \to \infty$, where the context is the entire host molecule. While explicit ECFP binning is impossible at this limit due to data sparsity, a language model natively attends to the full molecular sequence. This directly motivates our use of an LLM for Stage~1 fragment ranking, as it implicitly captures this global context without being bottlenecked by discrete cell fragmentation.

\begin{table}[!ht]
\caption{Variance reduction from ECFP2 grouping fragment attributions by local environment, compared against random shuffled grouping. Top K refers to the K most frequently occurring fragments in the dataset. The gap $\Delta=\mathrm{real}-\mathrm{shuf}$ isolates genuine context dependence.}
\label{tab:vr-main}
\centering
\small
\resizebox{\columnwidth}{!}{
\begin{tabular}{lccccc}
\toprule
Model & Top K & Real VR & Shuf. VR & $\Delta$ & Real$>$Shuf. \\
\midrule
Uni-Mol & 30  & 15.82\% & 7.59\%  & 8.23\% & 96.7\% \\
Uni-Mol & 100 & 21.05\% & 12.43\% & 8.62\% & 89.9\% \\
Uni-Mol & 300 & 28.29\% & 19.08\% & 9.20\% & 77.7\% \\
\midrule
ChemBERTa & 30  & 17.28\% & 8.86\%  & 8.42\% & 99.4\% \\
ChemBERTa & 100 & 22.79\% & 13.92\% & 8.88\% & 91.4\% \\
ChemBERTa & 300 & 29.71\% & 19.84\% & 9.87\% & 79.4\% \\
\bottomrule
\end{tabular}
}
\end{table}

\begin{table}[!htp]
\caption{Effect of ECFP radius ($r$) on fragment decoupling for ChemBERTa (top-100 fragments per property).  VR and the gap $\Delta$ are defined as in Table~\ref{tab:vr-main}. \textit{Nfrags} denotes the total number of fragments, and \textit{Avg.\ envs} is the average number of unique local environments per fragment.}
\label{tab:vr-radius}
\centering
\small
\setlength{\tabcolsep}{4pt}
\begin{tabular}{cccccc}
\toprule
$r$ & Nfrags & Avg.\ envs & Real VR & Shuf. VR & $\Delta$ \\
\midrule
1 & 480 & 7.5   & 4.57\%  & 2.54\%  & 2.03\% \\
2 & 594 & 104.1 & 22.79\% & 13.92\% & 8.88\% \\
3 & 600 & 322.6 & 37.87\% & 24.63\% & 13.23\%\\
4 & 600 & 493.5 & 48.54\% & 31.62\% & 16.91\%\\
\bottomrule
\end{tabular}
\end{table}

\begin{figure*}[!t]
\centering
\includegraphics[width=\textwidth]{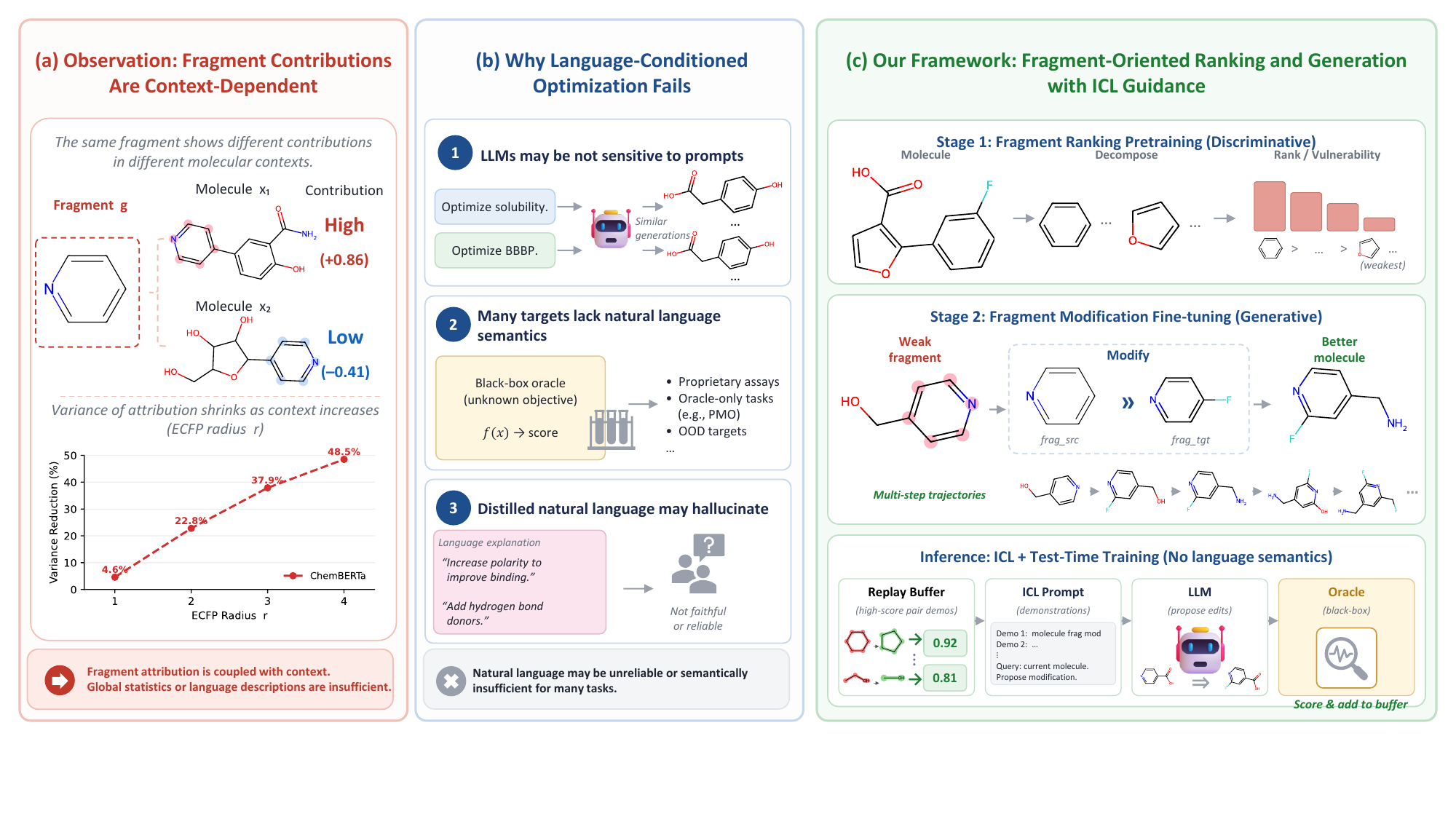}
\vspace{-1.6cm}
\caption{\textbf{Overview of the FORGE framework. (a)} Left: Empirical observations reveal that fragment contributions are highly context-dependent. \textbf{(b)} Middle: Natural language is a suboptimal interface for optimization due to weak prompt control, lack of oracle semantics, and distillation hallucinations. \textbf{(c)} Right: The FORGE pipeline decomposes the task into context-aware fragment ranking (Stage 1) and explicit modification (Stage 2), guided by dynamic in-context learning (ICL) at inference.}
\label{fig:framework}
\vspace{-0.5cm}
\end{figure*}

\subsection{What these observations imply for model design}
The evidence above points to a misalignment in current chemistry LLMs: prompt semantics are not enough to steer black-box optimization, and context-free fragment statistics discard the structural dependencies revealed by our VR analysis. Fragments themselves, however, are natural units of chemical knowledge—functional groups, scaffolds, and matched-pair edits familiar from medicinal chemistry. Therefore, the remedy is not to abandon fragment-level supervision, but to preserve its host-molecule context. We thus reformulate molecular optimization as \emph{context-aware local editing}.

From this, two principles follow: a model should derive its target semantics dynamically from oracle feedback rather than static text prompts, and it should operate in an explicit, context-conditioned edit space rather than re-decoding the whole molecule.

\section{Method: FORGE}
\label{sec:method}

FORGE (\textbf{F}ragment-\textbf{O}riented \textbf{R}anking and
\textbf{GE}neration) realises the context-aware local-editing view from
Section~\ref{sec:motivation} as a concrete training and inference
pipeline. The model is trained in two supervised stages: Stage~1 ranks
fragments under full-molecule context to learn \emph{where}, and Stage~2 generates an explicit
replacement for a chosen fragment to learn \emph{how}. An atom-level retokenization scheme
keeps fragment identity stable across host molecules so that this
fragment-level supervision is well defined. At test time, the same model
adapts to unnamed black-box objectives by conditioning on demonstrations
sampled from a replay buffer of (edit, score) tuples. We describe each
component below; Figure~\ref{fig:framework} gives an overview. Stage 1 instantiates context-aware where; Stage 2 instantiates context-aware how via SME+ and MMPA; the inference loop instantiates objective-from-feedback.

Both Stage~1 and Stage~2 are trained via supervised next-token prediction on \texttt{(instruction, input, output)} triples, computing the loss exclusively on the output span. Across both stages, we use four sources of supervision: GNN attribution, RDKit attribution, SME+, and ChEMBL matched molecular pairs. A detailed summary of the dataset construction is provided in Appendix~\ref{app:supervision}.

\subsection{Stage~1: learning where to edit}
\label{sec:stage1}

Stage~1 is supervised by fragment attribution. For properties with experimental labels but no closed-form evaluator, such as hERG, mutagenicity, lipop, BBBP and ESOL, we use RGCN models with mask-aware readout from SME and define fragment contribution by score drop after masking the fragment. For rule-based properties with exact evaluators, such as QED, logP, SA, and TPSA, we remove the fragment explicitly and recompute the property with RDKit. The first source provides broad learned supervision; the second provides low-noise labels that are exact under the metric.

From these attributions, we build four task families: \emph{decomposition} (split a molecule into atom-mapped fragments), \emph{ranking} (sort fragments by contribution to a target property), \emph{vulnerability prediction} (return the single weakest fragment), and \emph{ICL ranking/vulnerability} (perform the same outputs from in-context demonstrations, with or without an explicit property name). We also mix in Mol-Instructions \citep{2023arXiv230608018F} to keep general chemistry-language ability from being washed out by structured supervision. The full Stage~1 corpus contains about 2M examples; detailed mixtures are given in Appendix~\ref{app:data}.

This stage makes the edit decision context-aware and injects chemical prior. The model always reads the full molecular sequence, but the supervision remains fragment-level. As a result, Stage~1 does not learn a single global value for a fragment; it learns to judge that fragment in the context of the host molecule and ICL examples.

\subsection{Stage~2: learning how to edit}
\label{sec:stage2}

Stage~2 is trained on verified fragment replacement pairs. Its purpose is to map an identified weak region to a concrete local modification.

The first source of Stage~2 supervision is \textbf{SME+}, a context-conditioned extension of SME. Standard SME aggregates all observed effects of a fragment into a global range, ignoring the context dependence shown in Section~\ref{sec:motivation}. SME+ instead groups fragment occurrences by their local attachment environment, using the ECFP radius-2 neighborhood around the attachment atoms, and searches for replacements that improve the score within that context bin. When an environment bin is too sparse, SME+ falls back to the global pool. All proposed replacements are filtered by atom-map consistency checks, and only verified low-to-high pairs are retained. These pairs are also chained into multi-step trajectories, which later serve as in-context demonstrations during inference.

The second source is matched molecular pairs from ChEMBL \citep{10.1093/nar/gkr777}. For real-world targets, where no rule-based evaluator exist, so we rely on measured \texttt{pchembl}. We extract pairs that share a scaffold, differ at one fragment, and are measured on the same target. Direction is assigned by activity improvement, so each pair defines a low-to-high edit. To avoid leakage into downstream benchmarks, we exclude all pairs whose target description matches DRD2, JNK3, or GSK3-$\beta$.

Stage~2 uses the same output template for all sources:
\begin{verbatim}
Modification: <frag_src> >> <frag_tgt>
Result: <tgt_smi> [(value: s)]
\end{verbatim}
All SMILES spans are wrapped in \texttt{<start\_smiles>...\allowbreak<end\_smiles>} so that atom-level tokenization is applied consistently. The score suffix is included in most samples and omitted in the rest, so the model remains robust to whether oracle values are available at inference.

The backbone Stage~2 data merges SME+ trajectories on five ADMET properties (ESOL/ BBBP/ hERG/ lipop/ Mutag) with ChEMBL matched molecular pairs. The resulting checkpoint after Stage~1 and this backbone Stage~2 training is the \textbf{backbone} used for all downstream evaluations. Some benchmarks allow a small amount of additional task-specific continuation in the same format; those benchmark-specific details are described in Section~\ref{sec:experiments}.

\subsection{Atom-level retokenization}
\label{sec:qwenatom}

Fragment-level supervision only works if the same fragment is represented consistently across molecules. Standard BPE tokenization does not guarantee this: the same SMILES substring may be split differently depending on the surrounding sequence, which breaks the correspondence between fragments and model tokens.

We therefore augment the Qwen3 vocabulary with a small atom-level SMILES lexicon, including \texttt{<start\_smiles>}, \texttt{<end\_smiles>}, and \texttt{<smi>}-prefixed tokens for atoms in the organic subset, aromatic atoms, two-letter elements, bonds, brackets, and ring digits. Inside marked SMILES spans, tokenization switches to a regex-based atomic parser. The same fragment is therefore tokenized identically regardless of its host molecule.

The new embeddings are warm-started from their literal counterparts in the original vocabulary, and the LM head is initialized the same way. The first forward pass, therefore, matches the original Qwen3 distribution, so training begins from a stable pretrained prior rather than a newly learned tokenizer. We name this design as QwenAtom. In Appendix~\ref{app:stage1}, we show that this retokenization improves all 13 held-out Stage~1 ranking sub-tasks over vanilla Qwen3 tokenization.

\subsection{Inference under black-box objectives}
\label{sec:inference}

We maintain a replay buffer containing tuples of source molecule, applied modification, resulting molecule, and oracle score. At each step, the model is prompted with a small set of demonstrations sampled from this buffer, biased toward higher-scoring examples but tempered to preserve diversity. It then generates candidate modifications for the current molecule, the oracle scores the resulting molecules, and the new tuples are written back into the buffer. The buffer thus acts as an external memory of which local edits have worked under the current objective.

The output grammar is the same in Stage~2 and inference. Demonstration-conditioned examples, including no-description settings, are already present during training. The inference problem has the same structure as the training decomposition: the model must infer which fragment to change from full-molecule context and demonstrations, then propose an explicit replacement for that fragment.

\section{Experiments}
\label{sec:experiments}

\subsection{Experimental setup}

All experiments use Qwen3-0.6B as the base model, together with the \emph{QwenAtom} tokenizer from Section~\ref{sec:qwenatom}. Unless otherwise stated, the evaluated model is the shared \textbf{backbone} obtained from Stage~1 and Stage~2 training. Training uses standard language-model supervision with DeepSpeed ZeRO-1; implementation details and hyperparameters are deferred to Appendix~\ref{app:hp}.

We evaluate FORGE on five benchmarks. Prompt-MolOpt~\citep{wu2024leveraging} for similarity-constrained ADMET optimization, PMO-1k~\citep{gao2022sample} for sample-efficient adaptation to 22 unnamed black-box objectives, QED-DRD2 to isolate the effect of explicit fragment modification under a dual-objective similarity constraint, Lead Optimization~\citep{2022arXiv220811126W} to test real-target problem and ChemCoTBench~\citep{li2025beyond} for comparison against larger text-only and latent-reasoning models on six optimization tasks.

We additionally verify the tokenizer design in a controlled Stage~1 comparison, reported in Appendix~\ref{app:stage1}. Holding data, optimization, and training steps fixed, atom-level tokenization improves all 13 held-out Stage~1 sub-tasks over vanilla tokenization, with the largest gains on ICL and vulnerability prediction. This supports the role of stable fragment identity in the fragment-ranking stage.

\subsection{Prompt-MolOpt: similarity-constrained task}
\label{sec:exp:prompt}

Prompt-MolOpt \citep{wu2024leveraging} requires the model to improve one of five ADMET properties while maintaining a minimum similarity to the source molecule. For this benchmark, we initialize with the core FORGE backbone (comprising QwenAtom, Stage 1 (S1), and Stage 2 (S2)) and perform a brief task-specific fine-tuning (FT) using multi-turn SME+ data. We compare it against two ablations: removing Stage~1, and removing all Stage~1, Stage~2 and the multi-turn fine-tuning (FT on SME+ data with end-to-end SMILES output).

\begin{table}[!htp]
\caption{Prompt-MolOpt performance at 3 similarity thresholds. Best results under each similarity constraint are highlighted in bold.
}
\label{tab:prompt-molopt}
\centering
\small
\setlength{\tabcolsep}{5pt}
\renewcommand{\arraystretch}{1.}
\resizebox{\columnwidth}{!}{
\begin{tabular}{@{}lccccccc@{}}
\toprule
Model & $\delta$ & ESOL & BBBP & hERG & lipop & Mutag & SUM$\uparrow$ \\
\midrule
\multicolumn{8}{@{}l}{FORGE (QwenAtom + S1 + S2) + Multi-turn FT} \\
& 0   & \textbf{0.934} & \textbf{0.969} & \textbf{0.863} & \textbf{0.993} & 0.828 & \textbf{4.587} \\
& 0.4 & 0.780 & \textbf{0.949} & \textbf{0.847} & \textbf{0.982} & \textbf{0.744} & \textbf{4.302} \\
& 0.6 & 0.376 & \textbf{0.715} & \textbf{0.575} & 0.832 & \textbf{0.384} & \textbf{2.882} \\
\midrule
\multicolumn{8}{@{}l}{QwenAtom + S2 + Multi-turn FT} \\
& 0   & 0.928 & 0.941 & 0.828 & 0.993 & 0.784 & 4.474 \\
& 0.4 & \textbf{0.807} & 0.915 & 0.811 & 0.979 & 0.723 & 4.235 \\
& 0.6 & \textbf{0.400} & 0.676 & 0.566 & \textbf{0.842} & 0.376 & 2.860 \\
\midrule
\multicolumn{8}{@{}l}{QwenAtom + Single-turn FT} \\
& 0   & 0.892 & 0.918 & 0.803 & 0.991 & 0.775 & 4.379 \\
& 0.4 & 0.750 & 0.890 & 0.787 & 0.980 & 0.707 & 4.114 \\
& 0.6 & 0.345 & 0.682 & 0.559 & 0.841 & 0.359 & 2.786 \\
\midrule
\multicolumn{8}{@{}l}{Prompt-MolOpt~\citep{wu2024leveraging}} \\
& 0   & 0.893 & 0.936 & 0.857 & 0.923 & \textbf{0.857} & 4.466 \\
& 0.4 & 0.443 & 0.527 & 0.666 & 0.674 & 0.400 & 2.710 \\
& 0.6 & 0.104 & 0.164 & 0.257 & 0.225 & 0.074 & 0.824 \\
\bottomrule
\end{tabular}
}
\end{table}

Table~\ref{tab:prompt-molopt} shows that the gap between FORGE and Prompt-MolOpt grows as the similarity constraint tightens. At $\delta=0$, the gain over Prompt-MolOpt is small; at higher thresholds, the gap widens sharply because FORGE degrades much more gracefully. This is the setting where local fragment editing is most useful: staying close to the source is easier through targeted replacements than through full-SMILES rewriting.

The ablations decompose this gain. Removing Stage~2 and the multi-turn continuation hurts performance at all thresholds, indicating that explicit modification is the main source of improvement under similarity constraints. Removing Stage~1 also lowers performance, though more mildly, suggesting that fragment ranking mainly helps identify where to edit. We additionally find that the multi-turn formulation is most useful on longer edit trajectories, where carrying previous edits in context helps maintain a coherent local search direction; a direct comparison against repeated single-turn calls is reported in Appendix~\ref{app:pmo}.

\subsection{PMO-1k: limited oracle call black-box optimization}
\label{sec:exp:pmo}

PMO \citep{gao2022sample} directly tests the setting from Section~\ref{sec:motivation}: the objective is black-box, often unnamed, and accessible only through oracle calls. We evaluate under a reduced 1000-call budget rather than the standard 10k budget following LICO~\citep{nguyen2024lico}, which emphasizes sample efficiency. No task-specific continuation is used here; all adaptation comes from the replay-buffer ICL mechanism from Section~\ref{sec:inference}.

\begin{table}[!htp]
\caption{PMO-1k aggregate score, summed over 22 tasks. Per-task results are reported in Appendix~\ref{app:pmo}.}
\label{tab:pmo}
\centering
\small
\resizebox{\columnwidth}{!}{
\begin{tabular}{lc}
\toprule
Method & 22-task SUM $\uparrow$ \\
\midrule
GP-BO~\citep{2009arXiv0912.3995S} & 11.27 \\
Graph GA~\citep{Jensen2019Graph} & 10.90 \\
REINVENT~\citep{2017arXiv170407555O} & 10.68 \\
LICO~\citep{nguyen2024lico} & 11.71 \\
Genetic GFN~\citep{kim2024genetic} & 11.56 \\
Augmented Memory~\citep{2023arXiv230516160G} & 10.81 \\
MOLLEO~\citep{2024arXiv240616976W} & 11.65 \\
\midrule
\textbf{FORGE + ICL} & \textbf{12.42} \\
\bottomrule
\end{tabular}
}
\end{table}

FORGE reaches an aggregate score of 12.42 in Table~\ref{tab:pmo}, outperforming LICO by 0.71. The gain is consistent with the motivation of the method: in PMO, adaptation must come from examples and feedback.

We highlight three tasks. JNK3 improves from the best baseline score of 0.409 to 0.611, suggesting that the fragment editing prior learned from leakage-filtered ChEMBL MMPs transfers to hard real-target optimization. Both \textit{median1} and \textit{median2} reach state of the art. \textit{scaffold\_hop} also benefits from the fragment-based formulation.

\subsection{QED-DRD2: explicit fragment modification}
\label{sec:exp:qed-drd2}

QED-DRD2 isolates a single design decision: whether the model should output a full rewritten molecule or an explicit fragment modification. We compare two variants that share the same backbone, training data, and loss. The only difference is the output grammar during fine-tuning: \emph{end2end} predicts the whole target SMILES, while \emph{modif.} predicts the fragment replacement format used by FORGE.

\begin{table}[!htp]
\caption{QED-DRD2 success rates at three similarity thresholds. (\% success rates)}
\label{tab:qed-drd2}
\centering
\small
\setlength{\tabcolsep}{5pt}
\begin{tabular}{lccc}
\toprule
Method & $\delta{=}0.4$ & $\delta{=}0.5$ & $\delta{=}0.6$ \\
\midrule
JTNN~\citep{jin2018junction}         & 8.62  & 5.88  & 2.62  \\
HierG2G~\citep{jin2020hierarchical}       & 14.50 & 7.88  & 4.12  \\
Modof~\citep{chen2021deep}         & 24.62 & 16.62 & 9.12  \\
Prompt-MolOpt~\citep{wu2024leveraging} & 29.00 & 25.12 & 17.87 \\
Modofm~\citep{chen2021deep}        & 38.88 & 27.25 & 13.00 \\
\midrule
FORGE + end2end FT         & 35.15 & 33.76 & 27.32 \\
\textbf{FORGE + modif. FT} & \textbf{40.00} & \textbf{36.76} & \textbf{30.41} \\
\bottomrule
\end{tabular}
\end{table}

Table~\ref{tab:qed-drd2} shows modification variant outperforms the end-to-end variant by 3.0--5.0 points and beats all prior methods at every threshold. The gap is largest at the strictest similarity threshold, where preserving most of the source structure matters most. This supports the central design choice of FORGE: under similarity constraints, explicit fragment replacement is a better output space than free-form generation.

\subsection{Goal-directed Lead Optimization on Docking Targets}
\label{sec:exp:lead-opt}

We test FORGE on real-target docking, the setting closest to drug-discovery practice. Following \citet{2022arXiv220811126W}, given a seed ligand, the task is to generate molecules with improved docking score subject to QED $\ge 0.6$, SA $\le 4$, and Tanimoto similarity $\ge \delta$ to the seed (Morgan fingerprints), with $\delta \in \{0.4, 0.6\}$. We use five protein targets (\textit{parp1}, \textit{fa7}, \textit{5ht1b}, \textit{braf}, \textit{jak2}) and three seeds per target drawn from known actives \citep{2022arXiv220607632L}, giving 30 tasks per $\delta$. We compare against Graph GA \citep{Jensen2019Graph} and GenMol\citep{lee2025genmol}. Docking is the only oracle; FORGE adapts through the inference-time replay buffer without per-target fine-tuning.

\begin{table}[!htp]
\caption{Lead optimization, per-seed docking scores (kcal/mol; lower is better). Constraints: QED $\ge 0.6$, SA $\le 4$, Tanimoto sim $\ge \delta$. ``--'' = run produced no molecule satisfying all constraints. \textbf{Bold} = best in its $\delta$ block.}
\label{tab:lead-opt}
\centering
\scriptsize
\setlength{\tabcolsep}{3pt}
\begin{tabular}{llcccccccc}
\toprule
& & & \multicolumn{3}{c}{$\delta=0.4$} & & \multicolumn{3}{c}{$\delta=0.6$} \\
\cmidrule(lr){4-6}\cmidrule(lr){8-10}
Target & Seed & & GenMol & GraphGA & FORGE & & GenMol & GraphGA & FORGE \\
\midrule
\multirow{3}{*}{parp1}
 & $-7.3$ & & -10.60 & -8.30 & \textbf{-12.07} & & -10.40 & -8.60 & \textbf{-11.27} \\
 & $-7.8$ & & -11.00 & -8.90 & \textbf{-11.90} & & -9.70  & -8.10 & \textbf{-10.93} \\
 & $-8.2$ & & -11.30 & --    & \textbf{-11.83} & & -9.20  & --    & \textbf{-10.70} \\
\midrule
\multirow{3}{*}{fa7}
 & $-6.4$ & & -8.40  & -7.80 & \textbf{-9.17}  & & -7.30  & -7.60 & \textbf{-8.20}  \\
 & $-6.7$ & & -8.40  & -8.20 & \textbf{-9.07}  & & -7.60  & -7.60 & \textbf{-7.93}  \\
 & $-8.5$ & & --     & --    & --  & & --     & --    & --  \\
\midrule
\multirow{3}{*}{5ht1b}
 & $-4.5$ & & \textbf{-12.90} & -11.70 & -12.07 & & -12.10 & -11.30 & \textbf{-13.37} \\
 & $-7.6$ & & \textbf{-12.30} & -12.10 & -11.37 & & \textbf{-12.00} & \textbf{-12.00} & -10.07 \\
 & $-9.8$ & & \textbf{-11.60} & --     & -11.07 & & \textbf{-10.50} & --     & --     \\
\midrule
\multirow{3}{*}{braf}
 & $-9.3$ & & -10.80 & -9.80  & \textbf{-11.10} & & --     & --     & \textbf{-9.70}  \\
 & $-9.4$ & & \textbf{-10.80} & --     & -10.17 & & \textbf{-9.70}  & --     & --     \\
 & $-9.8$ & & -10.60 & \textbf{-11.60} & -10.73 & & -10.50 & -10.40 & \textbf{-10.60} \\
\midrule
\multirow{3}{*}{jak2}
 & $-7.7$ & & -10.20 & -8.70  & \textbf{-10.60} & & -9.30  & -8.10 & \textbf{-10.03} \\
 & $-8.0$ & & -10.00 & -9.20  & \textbf{-10.90} & & -9.40  & -9.20 & \textbf{-10.53} \\
 & $-8.6$ & & -9.80  & --     & \textbf{-10.80} & & --     & --    & \textbf{-9.80}  \\
\midrule
\multicolumn{3}{l}{\# best of 15} & 4 & 1 & \textbf{9} & & 3 & 1 & \textbf{11} \\
\bottomrule
\end{tabular}
\end{table}

Table~\ref{tab:lead-opt} reports the per-seed docking score for every (target, seed, $\delta$) configuration. FORGE achieves the best score on 9/15 runs at $\delta=0.4$ and 11/15 at $\delta=0.6$, beating GenMol and Graph GA at both similarity floors. The advantage is most consistent on \textit{parp1}, \textit{fa7}, and \textit{jak2}, where FORGE wins all per-seed runs. Graph GA, which mutates without context, trails on most targets and frequently fails to satisfy the similarity floor at $\delta=0.6$.

The gap holds up under the tighter similarity floor. Comparing $\delta=0.4$ to $\delta=0.6$, GenMol's score on \textit{parp1} (seed $-7.8$) degrades from $-11.00$ to $-9.70$, while FORGE degrades from $-11.90$ to $-10.93$ and stays ahead. A similar pattern holds on \textit{jak2} and \textit{fa7} for every seed. This is consistent with the analysis in Section~\ref{sec:ana:smep}: context-conditioned fragment supervision is most useful when the optimizer cannot leave the neighborhood of the seed.

The are two cases where FORGE underperforms Genmol. On \textit{5ht1b}, GenMol holds a $0.2$--$1.9$\,kcal/mol advantage. Due to small initial pool, the performance of FORGE is not very stable. On \textit{braf} the three methods are within $1$\,kcal/mol of each other across seeds, with no clear winner. We expect both gaps to close with a larger FORGE backbone (cf.\ Section~\ref{sec:ana:stages}).

\subsection{ChemCoTBench: small FORGE versus larger language models}
\label{sec:exp:cotbench}

\begin{table*}[!htp]
\caption{ChemCoTBench results. Each cell reports property improvement $\Delta$ / success rate SR (\%).}
\label{tab:cotbench}
\centering
\small
\begin{tabular}{llcccccc}
\toprule
Family & Method & LogP & Solub. & QED & DRD2 & JNK3 & GSK3-$\beta$ \\
\midrule
\multirow{3}{*}{Text-only}
 & Qwen-3-8B            & 0.00 / 3 & 0.00 / 4 & 0.00 / 4 & 0.00 / 4 & -0.01 / 0 & 0.00 / 2 \\
 & Qwen-3-8B (SFT)      & 0.15 / 47 & 0.48 / 52 & 0.10 / 48 & 0.04 / 38 & -0.02 / 20 & 0.02 / 36 \\
 & GPT-4o               & -0.01 / 37 & 0.92 / 80 & 0.13 / 70 & 0.07 / 48 & -0.02 / 30 & 0.00 / 39 \\
\midrule
\multirow{2}{*}{Latent}
 & Coconut-Chem~\citep{ye2026latentchem} & 0.17 / 44 & 0.57 / 58 & 0.15 / 67 & 0.04 / 45 & 0.00 / 35 & 0.06 / 47 \\
 & LatentChem~\citep{ye2026latentchem}   & \textbf{1.37} / \textbf{96} & \textbf{1.53} / 89 & 0.18 / 83 & 0.26 / 74 & 0.08 / 60 & 0.17 / 82 \\
\midrule
\multirow{3}{*}{Our (0.6B)}
 & QwenAtom + FT      & 1.07 / 90 & 0.81 / 82 & 0.18 / 85 & 0.32 / 82 & 0.13 / 67 & 0.28 / 82 \\
 & QwenAtom + S2 + FT & 1.02 / 93 & 1.04 / \textbf{95} & 0.19 / 88 & 0.36 / 87 & 0.24 / 83 & 0.33 / 89 \\
 & \textbf{FORGE (QwenAtom + S1 + S2) + FT} & 1.06 / 94 & 1.03 / \textbf{95} & \textbf{0.25} / \textbf{93} & \textbf{0.37} / \textbf{91} & \textbf{0.27} / \textbf{86} & \textbf{0.34} / \textbf{94} \\
\bottomrule
\end{tabular}
\end{table*}

ChemCoTBench \citep{li2025beyond} supplies a 5k-pair training set and six optimization tasks: LogP, Solubility, QED, DRD2, JNK3, and GSK3-$\beta$. We extract fragment-replacement pairs from the official training set and fine-tuning the FORGE backbone in the same modification format, without external data, oracle calls, or chain-of-thought distillation. This matches the FORGE supervision to what is available to the text-only baselines.

Table~\ref{tab:cotbench} summarizes the comparative evaluation. At 0.6B parameters, FORGE (QwenAtom + Stage 1 + Stage 2) + FT matches or outperforms Qwen-3-8B (SFT) and GPT-4o on every task, with the largest gaps on the real-target tasks: on DRD2, JNK3, and GSK3-$\beta$. FORGE reaches success rates of $91\%$, $86\%$, and $94\%$, respectively. In contrast, Qwen-3-8B achieves only $38\%$, $20\%$, and $36\%$, while GPT-4o reaches $48\%$, $30\%$, and $39\%$. The property-improvement metric ($\Delta$) exhibits the same trend: Qwen-3-8B stays below $0.10$ on LogP, QED, DRD2, and GSK3-$\beta$, and even turns negative on JNK3. This indicates that the natural-language CoT format used in its training carries little optimization signal beyond improving basic output validity. Against the latent-reasoning baseline LatentChem, FORGE outperforms it on four out of six tasks across both metrics; while LatentChem retains a higher $\Delta$ on LogP and Solubility, FORGE matches or exceeds its overall success rate on all six tasks.

Notably, even the simplest ablation variant (QwenAtom + FT), which only fine-tunes the backbone on basic fragment replacements, still outperforms LatentChem and the 8B SFT baseline on several tasks. Given that these baselines rely on complex reinforcement learning or extensive chain-of-thought reasoning, this result further supports our argument: natural language is likely a suboptimal interface for molecular optimization.

\section{Analysis}
\label{sec:analysis}

The experiments show that FORGE performs well across similarity-constrained optimization, black-box adaptation, and real-target editing. Two questions remain after the main results: how much of the Prompt-MolOpt gain comes from data construction versus output format, and how do Stage~1 and Stage~2 divide labor? We now analyze them in turn.

\subsection{Context-conditioned data matters most under local-edit constraints}
\label{sec:ana:smep}

We isolate one design choice: building edit supervision from context-conditioned fragment effects (SME+) versus a single global statistic per fragment (the SME-style aggregation used by the original Prompt-MolOpt data). We compare two models with the same backbone and the same end-to-end output format, differing only in the training data used for Prompt-MolOpt continuation.

\begin{table}[!htp]
\caption{Effect of training-data aggregation on Prompt-MolOpt SUM. The backbone and output format are fixed.}
\label{tab:sme-vs-smeplus}
\centering
\small
\begin{tabular}{lccc}
\toprule
$\delta$ & SME+ & original SME-style & $\Delta$ \\
\midrule
0.0 & 3.781 & 3.977 & $-0.196$ \\
0.4 & 3.582 & 2.928 & $+0.654$ \\
0.6 & 2.187 & 1.354 & $+0.833$ \\
\bottomrule
\end{tabular}
\end{table}

The pattern in Table~\ref{tab:sme-vs-smeplus} flips with the similarity threshold. At $\delta=0$, the model is free to move to distant high-scoring regions, and global aggregation is slightly more useful. Once the similarity floor becomes difficult, SME+ wins by $+0.65$ SUM at $\delta=0.4$ and $+0.83$ at $\delta=0.6$. Within-context supervision matches the local-edit regime that the similarity floor enforces.

Switching to SME+ data alone adds $+0.833$ SUM at $\delta=0.6$, the single largest of the three factors in Table~\ref{tab:decomp}. Output interface and supervision data contribute on a comparable scale; the data construction is the single largest factor. This shows that FORGE benefits not only from a better output interface, but also from supervision that matches the local nature of similarity-constrained editing.

\begin{table}[!htp]
\caption{Decomposition of the gain over the original Prompt-MolOpt baseline at $\delta{=}0.6$.}
\label{tab:decomp}
\centering
\small
\begin{tabular}{lcc}
\toprule
Source & SUM $\delta{=}0.6$ & gain \\
\midrule
Prompt-MolOpt baseline & 0.824 & --- \\
$+$ backbone upgrade & 1.354 & $+0.530$ \\
$+$ SME+ data & 2.187 & $+0.833$ \\
$+$ modification format (S1+S2) & 2.882 & $+0.695$ \\
\bottomrule
\end{tabular}
\end{table}

\subsection{Stage~2 provides the main editing capability, while Stage~1 improves localization}
\label{sec:ana:stages}

\begin{figure}[!htp]
\centering
\includegraphics[width=\columnwidth]{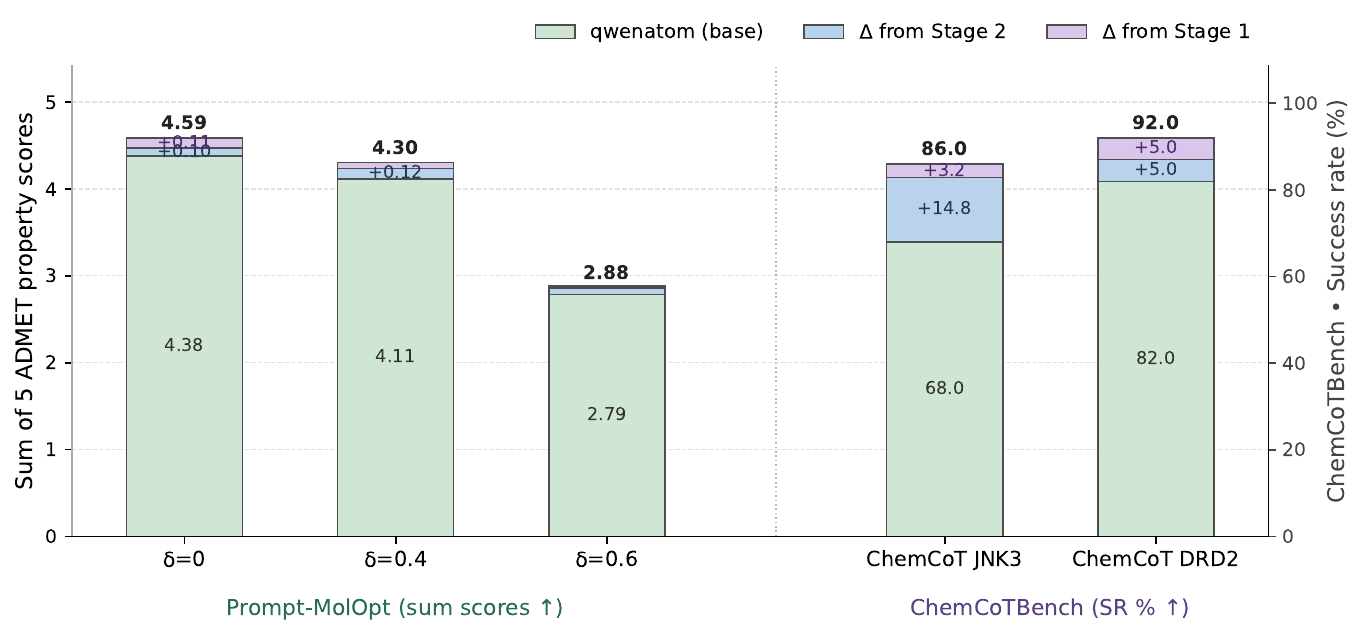}
\vskip 5pt
\caption{Marginal gains of Stage~1 and Stage~2 across representative benchmarks.}
\label{fig:stage-marginal}
\end{figure}

The ablations in Prompt-MolOpt and ChemCoTBench suggest Stage~1 and Stage~2 contribute differently across benchmarks. Table~\ref{tab:marginal} and Figure~\ref{fig:stage-marginal} summarize that pattern across representative settings.

\begin{table}[!htp]
\caption{Marginal gains of Stage~1 and 2 across settings.}
\label{tab:marginal}
\centering
\small
\begin{tabular}{lccc}
\toprule
Setting & qwenatom & $+$ S2 & $+$ S1+S2 \\
\midrule
Prompt-MolOpt $\delta{=}0$   & 4.379 & 4.474 & 4.587 \\
Prompt-MolOpt $\delta{=}0.4$ & 4.114 & 4.235 & 4.302 \\
Prompt-MolOpt $\delta{=}0.6$ & 2.786 & 2.860 & 2.882 \\
ChemCoT JNK3 $\Delta$        & 0.13  & 0.24  & 0.27 \\
ChemCoT DRD2 $\Delta$        & 0.32  & 0.36  & 0.37 \\
ChemCoT GSK3-$\beta$ $\Delta$ & 0.28 & 0.33  & 0.34 \\
\bottomrule
\end{tabular}
\end{table}

Stage~2 is the larger increment throughout. This is expected: Stage~2 supervision is in the same output space used at inference, so the gradient signal directly trains the inference behavior. The increment is largest on the real-target tasks in ChemCoTBench (JNK3 SR: $67\to83$; Solubility: $82\to95$).

Stage~1 adds a smaller but consistent gain on top of Stage~2. The improvement is most visible when the search space is broad or the oracle signal is sparse, such as unconstrained or moderately constrained Prompt-MolOpt and the real-target tasks in ChemCoTBench. Stage~1 therefore acts as a localisation prior on top of Stage~2 rather than as a substitute.

These results clarify why fragment-aware supervision fits molecular optimization better than property prediction. In standard property prediction, each molecule is paired with a single scalar label, so fragment-level credit assignment must be inferred indirectly. In optimization, by contrast, a source molecule, a local modification, and the resulting oracle change already provide supervision at roughly the fragment level. FORGE turns this structure into direct supervision via Stage~1 ranking and Stage~2 replacement.

FORGE does not dominate every PMO task. The per-task breakdown in Appendix~\ref{app:pmo} shows that the largest deficits occur against MOLLEO on tasks tied to specific named drugs or heavily studied targets, such as \textit{albuterol\_similarity}, \textit{perindopril\_mpo}, \textit{drd2}, and \textit{amlodipine\_mpo}; MOLLEO is the stronger baseline on these specific tasks even though LICO has the higher aggregate score. These tasks likely reward broad memorisation of named-drug chemistry, which favours much intensively pretrained backbones such as the BioT5~\citep{pei2023biot5} generators inside MOLLEO. We leave drug/target-specific demonstration pools to future work.

\subsection{Limitations}
\label{sec:limitations}
First, FORGE does not explicitly model synthesizability: SME+ and ChEMBL MMPs contain only realized molecules, but provide no supervision on whether a proposed edit matches a known reaction, which occasionally leads to outputs with high SA scores (e.g., in ChemCoTBench Solubility). Second, some failures are due to out-of-distribution property knowledge: the 0.6B base lacks large-scale memorized drug--structure associations, so these errors reflect a missing prior rather than an optimization failure and are unlikely to be fixed by ICL evolution alone. Third, Stages 1--2 do not span all properties: the backbone is trained on five ADMET properties and ChEMBL MMPs, making it a strong initialization but not a zero-shot generator; novel objectives still require the per-evaluation SFT step.

\section{Discussion and Conclusion}
\label{sec:conclusion}

FORGE casts molecular optimization as context-aware local editing on a 0.6B language model: atom-level tokenization keeps fragment identity stable across host molecules, Stage~1 ranks fragments under full-molecule context, Stage~2 generates explicit replacements supervised by SME+ pairs and ChEMBL MMPs, and inference adapts to unnamed objectives through a replay buffer of (edit, score) demonstrations. This combination exceeds Qwen3-8B (SFT) and GPT-4o on ChemCoTBench at 0.6B parameters, and sets the best aggregate scores we are aware of on Prompt-MolOpt (SUM $4.59/4.30/2.88$ at $\delta=0/0.4/0.6$), PMO-1k (SUM $12.42$ over 22 tasks), QED-DRD2, and Lead Optimization.

Two points generalize beyond this method. First, for fragment-based molecular optimization, the value of a fragment depends on
the molecule that hosts it, so within-context supervision (SME+) is 
worth more than the output-format change (Section~\ref{sec:ana:smep}). Second, when a black-box objective lacks a 
faithful natural-language description, demonstrations and oracle scores 
are a stronger control signal than prompt rewriting. Both observations 
are independent of model scale, and we expect the same pipeline to 
benefit from a larger backbone as named-drug priors become a bottleneck 
(Section~\ref{sec:ana:stages}). Two natural extensions are coupling the 
inference buffer with a retrosynthesis filter to address the SA-score 
issue identified in Section~\ref{sec:limitations}, and porting the 
fragment-replacement supervision format to reaction-prediction tasks 
where matched-pair data is also abundant.

\clearpage
\bibliography{ref}
\bibliographystyle{icml2026}

\newpage

\clearpage
\appendix

\section{Per-property Variance Reduction}
\label{app:vr}

\Cref{tab:vr-per-property} reports the per-property VR for the three properties shared across all three predictors in \cref{sec:vr}. We test the models on 10K mol pool of ZINC dataset. Uni-Mol and ChemBERTa are trained on MoleculeNet. RGCN is taken from SME. 

\begin{table}[!ht]
\caption{Per-property variance reduction on the three properties shared by SME (RGCN), Uni-Mol, and ChemBERTa.}
\label{tab:vr-per-property}
\centering
\small
\begin{tabular}{lccc}
\toprule
Property & SME (RGCN) & Uni-Mol & ChemBERTa \\
\midrule
ESOL & 17.32\% & 18.18\% & 14.32\% \\
Lipophilicity & 19.01\% & 20.19\% & 15.12\% \\
BBBP & 14.29\% & 14.95\% & 19.76\% \\
\bottomrule
\end{tabular}
\end{table}

\begin{figure}[!htp]
\centering
\includegraphics[width=\columnwidth]{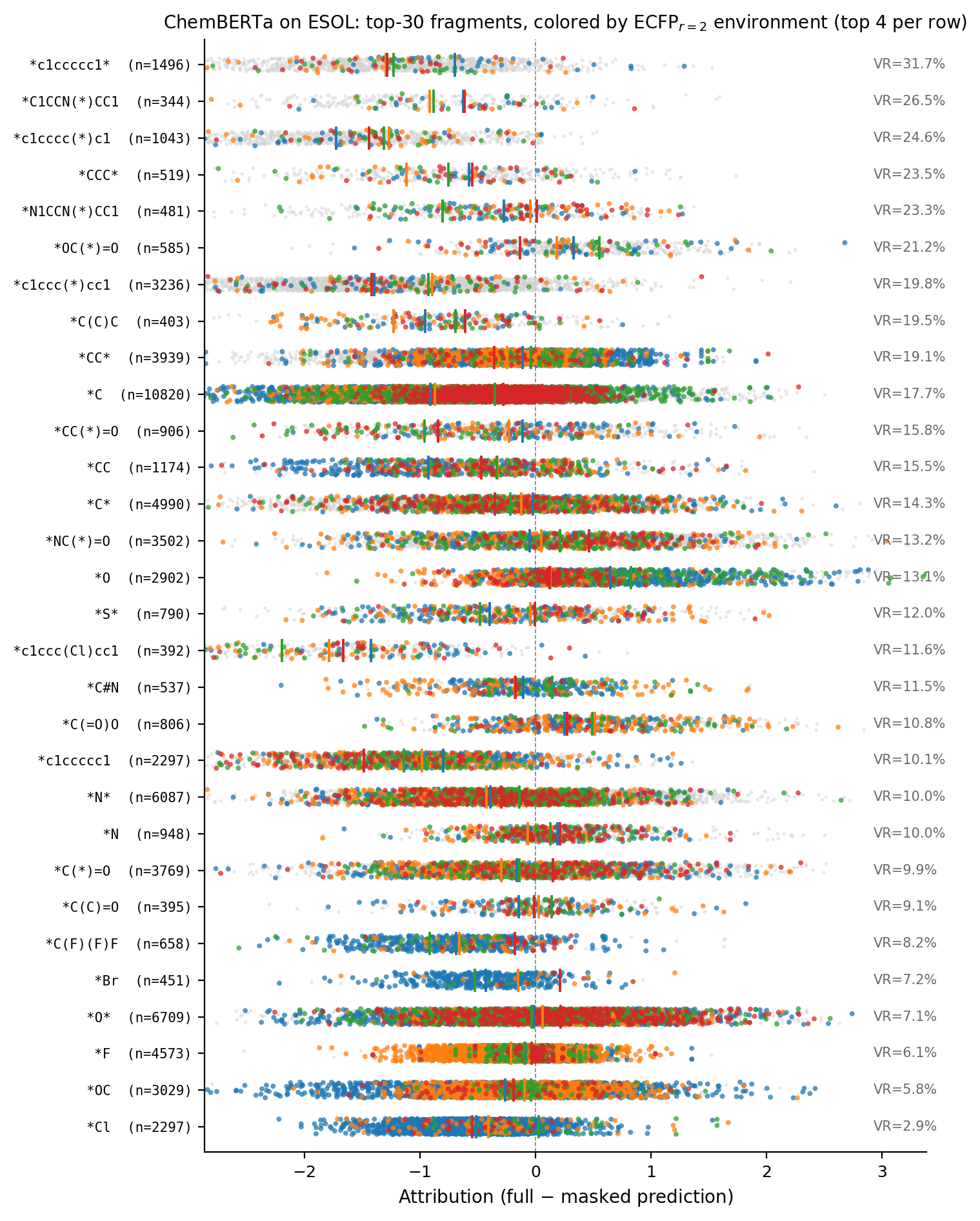}
\vskip 5pt
\caption{Fragment decoupling of ChemBERTa on lipop}
\label{fig:fig_fragment_strip}
\end{figure}

\section{More experiment details}
\label{app:pmo}

\Cref{tab:pmo-full} reports all 22 PMO tasks under the 1k-call budget. Columns: GP-BO~\citep{2009arXiv0912.3995S}, Graph GA~\citep{Jensen2019Graph}, REINVENT~\citep{2017arXiv170407555O}, LICO~\citep{nguyen2024lico}, Genetic GFN~\citep{kim2024genetic}, Augmented Memory~\citep{2023arXiv230516160G}, MOLLEO~\citep{2024arXiv240616976W}, FORGE backbone (\textit{qwenatom + S1 + S2}), and the FORGE Stage-1-ablation (\textit{qwenatom + S2}). Means across 5 runs are reported. The four tasks where MOLLEO outperforms FORGE — \textit{albuterol\_similarity}, \textit{amlodipine\_mpo}, \textit{drd2}, \textit{perindopril\_mpo} — are the named-drug-recall failure modes.

\begin{table*}[!ht]
\caption{PMO-1k per-task scores. The aggregate sum is reported in the bottom row; FORGE backbone reaches 12.42, $+0.71$ over LICO and $+0.77$ over MOLLEO. Best per row in bold.}
\label{tab:pmo-full}
\centering
\small
\setlength{\tabcolsep}{3pt}
\resizebox{\textwidth}{!}{
\begin{tabular}{lccccccccc}
\toprule
Task & GP-BO & Graph GA & REINVENT & LICO & Genetic GFN & Aug. Mem. & MOLLEO & \textbf{FORGE (S1+S2)} & FORGE (S2) \\
\midrule
albuterol\_similarity     & 0.636 & 0.583 & 0.496 & 0.656 & 0.664 & 0.557 & \textbf{0.886} & 0.779 & 0.764 \\
amlodipine\_mpo           & 0.519 & 0.501 & 0.472 & 0.541 & 0.534 & 0.489 & \textbf{0.637} & 0.551 & 0.554 \\
celecoxib\_rediscovery    & 0.411 & 0.424 & 0.370 & \textbf{0.447} & \textbf{0.447} & 0.385 & 0.402 & 0.428 & 0.441 \\
deco\_hop                 & 0.593 & 0.581 & 0.572 & 0.596 & 0.604 & 0.579 & 0.588 & \textbf{0.624} & \textbf{0.624} \\
drd2                      & 0.857 & 0.833 & 0.775 & 0.859 & 0.809 & 0.795 & \textbf{0.910} & 0.816 & 0.818 \\
fexofenadine\_mpo         & 0.707 & 0.666 & 0.650 & 0.700 & 0.682 & 0.679 & 0.674 & 0.740 & \textbf{0.742} \\
gsk3b                     & 0.611 & 0.523 & 0.589 & 0.617 & \textbf{0.637} & 0.539 & 0.397 & 0.620 & 0.581 \\
isomers\_c7h8n2o2         & 0.545 & 0.735 & 0.725 & \textbf{0.779} & 0.738 & 0.661 & 0.737 & 0.723 & 0.652 \\
isomers\_c9h10n2o2pf2cl   & 0.599 & 0.630 & 0.630 & 0.672 & 0.656 & 0.596 & 0.635 & \textbf{0.713} & 0.712 \\
jnk3                      & 0.346 & 0.301 & 0.315 & 0.336 & 0.409 & 0.294 & 0.186 & \textbf{0.611} & 0.584 \\
median1                   & 0.213 & 0.208 & 0.205 & 0.217 & 0.219 & 0.219 & 0.236 & \textbf{0.271} & 0.250 \\
median2                   & 0.203 & 0.181 & 0.188 & 0.193 & 0.204 & 0.184 & 0.191 & \textbf{0.221} & 0.220 \\
mestranol\_similarity     & 0.427 & 0.362 & 0.379 & 0.423 & 0.414 & 0.393 & 0.399 & 0.450 & \textbf{0.453} \\
osimertinib\_mpo          & 0.766 & 0.751 & 0.737 & 0.759 & 0.763 & 0.761 & 0.779 & \textbf{0.794} & 0.787 \\
perindopril\_mpo          & 0.458 & 0.435 & 0.404 & 0.473 & 0.462 & 0.422 & \textbf{0.655} & 0.509 & 0.522 \\
qed                       & 0.912 & 0.914 & 0.921 & 0.925 & 0.928 & 0.923 & 0.919 & \textbf{0.937} & \textbf{0.937} \\
ranolazine\_mpo           & 0.701 & 0.620 & 0.574 & 0.687 & 0.623 & 0.614 & 0.640 & 0.680 & 0.690 \\
scaffold\_hop             & 0.478 & 0.461 & 0.447 & 0.480 & 0.485 & 0.460 & 0.473 & \textbf{0.532} & 0.511 \\
sitagliptin\_mpo          & 0.232 & 0.229 & 0.261 & 0.315 & 0.227 & 0.245 & 0.193 & 0.300 & \textbf{0.337} \\
thiothixene\_rediscovery  & 0.351 & 0.322 & 0.311 & 0.343 & 0.377 & 0.336 & \textbf{0.416} & 0.400 & 0.374 \\
troglitazone\_rediscovery & \textbf{0.313} & 0.267 & 0.246 & 0.292 & 0.277 & 0.262 & 0.302 & 0.288 & 0.286 \\
valsartan\_smarts         & 0.000 & 0.000 & 0.000 & 0.000 & 0.000 & 0.000 & 0.000 & 0.000 & 0.000 \\
zaleplon\_mpo             & 0.392 & 0.374 & 0.406 & 0.404 & 0.400 & 0.415 & 0.392 & 0.435 & \textbf{0.436} \\
\midrule
SUM ($\uparrow$) & 11.27 & 10.90 & 10.68 & 11.71 & 11.56 & 10.81 & 11.65 & \textbf{12.42} & 12.27 \\
\bottomrule
\end{tabular}
}
\end{table*}

For Prompt-MolOpt, multi-turn generation and repeated single-turn generation are similar for two-step trajectories, but multi-turn becomes preferable as the number of edits increases as shown in \cref{tab:multi-vs-iter}. This is consistent with the role of dialogue context in preserving a coherent sequence of local modifications.

\begin{table}[!ht]
\caption{Similarity to the source molecule on Prompt-MolOpt, comparing a single multi-turn dialogue against repeated single-turn calls for trajectories of the same length. Higher is better.}
\label{tab:multi-vs-iter}
\centering
\small
\begin{tabular}{lccc}
\toprule
\# turns & multi-turn & iter & $n$ \\
\midrule
2 & 0.487 & \textbf{0.495} & 7274 \\
3 & \textbf{0.483} & 0.468 & 1164 \\
4 & \textbf{0.474} & 0.447 & 328 \\
\bottomrule
\end{tabular}
\end{table}

\section{Hyperparameters}
\label{app:hp}

All training runs use AdamW with peak learning rate $2\!\times\!10^{-5}$, linear warm-up over 5{,}000 steps and cosine decay, per-device batch size 4, gradient accumulation 1, sequence length 2048, bf16 mixed precision, and DeepSpeed ZeRO-1 across 4$\times$NVIDIA A100 (80GB) GPUs. Loss is the standard left-to-right token cross-entropy; for ICL samples and multi-turn answers, the loss is masked to the answer span only. Per-stage epoch counts are summarized in \cref{tab:hp}.

\begin{table}[!ht]
\centering
\caption{Per-stage epoch counts. All other hyperparameters (optimizer, lr, warm-up, batch size, hardware) are shared.}
\label{tab:hp}
\setlength{\tabcolsep}{4pt}
\small
\resizebox{\columnwidth}{!}{
\begin{tabular}{lc}
\toprule
Stage & Epochs \\
\midrule
Stage 1 (ranking + decomposition + ICL) & 3 \\
Stage 2 backbone (single-step modification) & 1 \\
Stage 2 multi-turn (Prompt-MolOpt) & 1 \\
Other downstream finetune (QED-DRD2, ChemCoTBench, etc.) & 3 \\
\bottomrule
\end{tabular}
}
\end{table}

For inference: vLLM with temperature $\tau{=}0.7$, top-$p{=}0.95$, max generation length 512 (single-step) or 1024 (multi-turn). The ICL replay buffer keeps the top 200 cache and samples with the top $K{=}5$ verified pairs by improvement $\Delta$ and is refreshed after every successful step.

\section{Data Construction Details}
\label{app:data}

This appendix documents the recipes used to build each training corpus referenced in \cref{sec:method}: source, volume, output schema, and a representative example. Final mixture statistics are summarized in \cref{tab:stage1mix,tab:stage2mix}.

\subsection{Training Data and Supervision Details}
\label{app:supervision}

Table~\ref{tab:data-roles-app} summarizes the supervision sources used across the two training stages of FORGE. Stage~1 uses fragment-level attribution labels to teach the model where to edit, while Stage~2 uses verified fragment replacement pairs to teach how to edit. Among these sources, ChEMBL matched molecular pairs support both stages: they provide demonstration-based ranking supervision in Stage~1 and explicit low-to-high edit pairs in Stage~2.

\begin{table}[ht]
\caption{Data sources and supervision across training stages.}
\label{tab:data-roles-app}
\centering
\small
\resizebox{\columnwidth}{!}{
\begin{tabular}{llc}
\toprule
Source & Supervision & Training Stage \\
\midrule
GNN mask attribution & ranking labels & 1 \\
RDKit rule attribution & ranking labels & 1 \\
SME+ & verified pairs & 2 \\
ChEMBL MMPA & verified pairs / ICL demonstrations & 1 \& 2 \\
\bottomrule
\end{tabular}
}
\end{table}

\paragraph{Stage~1 supervision.}
GNN-based attribution is used for properties without closed-form evaluators, including hERG, BBBP, mutagenicity, and solubility. RDKit-based attribution is used for rule-based properties, including QED, LogP, SA, and TPSA. Both sources produce fragment-level ranking and vulnerability labels.

\paragraph{Stage~2 supervision.}
SME+ produces context-conditioned low-to-high fragment replacement pairs on five ADMET properties: ESOL, BBBP, hERG, lipophilicity, and mutagenicity. ChEMBL matched molecular pairs provide real-activity fragment replacement pairs constructed from measured \texttt{pchembl} differences on the same target.

\paragraph{Leakage prevention for ChEMBL.}
To avoid overlap with downstream evaluations, we remove all ChEMBL pairs whose target description matches DRD2, JNK3, or GSK3-$\beta$. This filtering is applied before constructing the final training pool.

\subsection{Stage 1: discriminative fragment-level tasks (\textasciitilde 2M samples)}

Stage 1 trains the model to \emph{rank} fragments by their per-atom contribution to a property and to identify the weakest fragment (a vulnerability prompt). Sources and proportions are listed in \cref{tab:stage1mix}; smaller sources are oversampled to match their target ratio.

\begin{table}[!t]
\centering
\caption{Stage 1 source mixture, totalling 2M training examples.}
\label{tab:stage1mix}
\setlength{\tabcolsep}{4pt}
\small
\begin{tabular}{lc}
\toprule
Source & Ratio \\
\midrule
RDKit fragment attribution (ranking + vulnerability)   & 25\% \\
GNN mask attribution (ranking)                         & 15\% \\
GNN decomposition prediction                           & 10\% \\
ICL ranking (RDKit + ChEMBL MMP demonstrations)        & 35\% \\
Mol-Instructions (external instruction-following)      & 15\% \\
\bottomrule
\end{tabular}
\end{table}

\paragraph{RDKit fragment attribution.} We sample \textasciitilde 50k molecules per property from ZINC. Each molecule is decomposed by one of \{Murcko, BRICS, EFG\} (chosen in random order; the first decomposition that yields $\geq 3$ fragments wins). For every fragment, we recompute the molecule's property after two removal procedures, replace-with-H and delete-with-cap, and store the per-atom delta $\Delta/n_\text{atoms}$ to remove a known size bias. Twelve RDKit properties are covered: QED, logP, SA, TPSA, molecular weight, HBA, HBD, rotatable bonds, aromatic rings, Fsp3, ring count, and heavy-atom count. Each (molecule, property, removal) triple is converted into either a \emph{ranking} prompt (sort all fragments) or a \emph{vulnerability} prompt (return the weakest), and 35\% of prompts request the normalized contribution score in addition to the ordering. A representative example:

\begin{tcolorbox}[colback=gray!5!white, colframe=gray!70!black, arc=3pt, boxrule=0.5pt, left=4pt, right=4pt, top=4pt, bottom=4pt, halign=left] 
\small
\textbf{Instruction:} Which fragment contributes the most to 'number of rings'? Rank all fragments and include their normalized contribution scores (0 = lowest, 1 = highest observed contribution).

\vspace{0.5em}
\textbf{Input:} Molecule SMILES: \seqsplit{\textless start\_smiles\textgreater C=CCNC(=O)CN1CC(S(=O)(=O)N2CCNCC2)C1\textless end\_smiles\textgreater}
List of fragments to be ranked: \seqsplit{\textless start\_smiles\textgreater[3*]C[4*]. ...\textless end\_smiles\textgreater}

\vspace{0.5em}
\textbf{Output:} \seqsplit{\textless start\_smiles\textgreater[4*]N([5*])[10*]\textless end\_smiles\textgreater} (score: 0.14) \textgreater\ ... \textgreater\ \seqsplit{\textless start\_smiles\textgreater[9*]CC[11*]\textless end\_smiles\textgreater} (score: 0.0)
\end{tcolorbox}


\paragraph{GNN attribution.} A mirror of the RDKit pipeline that uses the trained SME RGCN as the property predictor (so it covers learned ADMET endpoints) and SME's substructure mask as the decomposition. Two task families are emitted: \emph{ranking} (sort fragments by attribution score) and \emph{decomposition} (predict the fragment list induced by the mask).

\paragraph{ICL ranking.} 100k RDKit-ICL and 500k ChEMBL-MMP-ICL samples. For RDKit-ICL, we draw $K{=}5$ demonstrations sharing the same (property, removal-method) cell from the attribution corpus and mask the property name, forcing pattern induction from demonstrations alone. For MMP-ICL, the demonstrations are different scaffolds against the same biological target, and the query is a held-out scaffold whose candidate fragments must be ranked by activity. We mix three prompt variants: ICL with target description (50\%), ICL without description (30\%), and direct ranking with no demonstrations (20\%); each variant switches to a vulnerability prompt with probability 0.25 and adds normalized scores with probability 0.35. Targets whose description mentions JNK, DRD, and GSK are excluded so they remain unseen at evaluation time.

\subsection{Stage 2: generative single-step modification (\textasciitilde 680k samples)}

Stage 2 teaches the model to emit a fragment-replacement edit and the resulting molecule. Sources and proportions are listed in \cref{tab:stage2mix}.

\begin{table}[!t]
\centering
\caption{Stage 2 source mixture (single-step backbone). The multi-turn variant for Prompt-MolOpt is trained as a separate continuation; see text.}
\label{tab:stage2mix}
\setlength{\tabcolsep}{4pt}
\small
\resizebox{\columnwidth}{!}{
\begin{tabular}{lcc}
\toprule
Source & Pairs (post-dedup) & Share \\
\midrule
SME+ verified ADMET pairs (5 properties)   & \textasciitilde 136k*2 & 40\% \\
ChEMBL MMP pairs (multi-target)            & \textasciitilde 411k & 60\% \\
\bottomrule
\end{tabular}
}
\end{table}

\paragraph{SME+ verified pairs.} Sub-trajectories whose atom mapping survives the SME+ verifier are flattened into single-step pairs covering five ADMET endpoints: ESOL, BBBP, hERG, lipophilicity, and mutagenicity. Each property is mapped through a sigmoid or sign flip so that ``higher is better'' holds uniformly, scores are min-max normalized per property to $[0,1]$, pairs with $\Delta_{\text{score}}\le 0.01$ are dropped, and each unique \emph{frag\_src $\rightarrow$ frag\_tgt} edit is capped at five occurrences to prevent a small set of common substitutions from dominating the loss. The single-step output template is

\begin{small}
\begin{verbatim}
Modification: <frag_src> >> <frag_tgt>
Result:       <tgt_smi> (value: 0.74)
\end{verbatim}
\end{small}

with the \texttt{(value:\,$\cdot$)} suffix present in 80\% of samples. Prompts are split between direct (10\%, naming the property by description or by name) and ICL (90\%, with $K{=}5$ demonstrations from the same property).

\paragraph{ChEMBL MMP optimization.} Matched-molecular pairs are mined across ChEMBL targets with pChEMBL $\geq 5$ on both sides. We orient each pair so that the higher-pChEMBL side is the optimization target, normalize pChEMBL per target to $[0,1]$, and discard any pair whose larger fragment exceeds 15 heavy atoms or covers more than 40\% of the parent (these tend to encode whole-scaffold swaps rather than local edits). The same \texttt{Modification / Result} template is used, but the score field is named \texttt{activity} instead of \texttt{value}. Prompts are mixed between direct (target named or described) and ICL (five demonstrations sharing the same target). DRD2, JNK3 and GSK3b are again held out by description match.

\paragraph{Multi-turn variant.} For Prompt-MolOpt, we additionally train a multi-turn continuation that reuses the verified SME+ trajectories but emits the entire sub-trajectory (capped at six steps) as a single answer, so the model learns to plan a sequence of edits in one shot. The training data statistics for each property are summarized as \cref{tab:multi-turn-stats}

\begin{table}[!ht]
\centering
\small
\caption{Statistics of multi-turn training trajectories across different properties.}
\begin{tabular}{lccc}
\toprule
\textbf{Property} & \textbf{Goal} & \textbf{Sub-trajs} & \textbf{Steps} \\
\midrule
lipop & Higher & 47,485 & 103,570 \\
ESOL & Higher & 33,191 & 72,295 \\
hERG & Lower & 31,247 & 63,413 \\
BBBP & Higher & 24,269 & 53,906 \\
Mutagenicity & Lower & 22,700 & 45,610 \\
\bottomrule
\end{tabular}
\label{tab:multi-turn-stats}
\end{table}

Properties are upsampled to the count of the largest property(lipop, 47485) to balance the mixture. The output is a concatenation of \texttt{Modification / Result} blocks:

\begin{small}
\begin{verbatim}
Modification: <frag_src_1> >> <frag_tgt_1>
Result: <mol_1> (value: 0.41)
Modification: <frag_src_2> >> <frag_tgt_2>
Result: <mol_2> (value: 0.58)
...
\end{verbatim}
\end{small}

\paragraph{Format normalization.} Across all stages, isotope brackets on atoms in the organic subset $\{B, C, N, O, P, S, F, Cl, Br, I\}$ are stripped to keep SMILES canonical, while dummy labels of the form $[k*]$ on fragment attachment points are preserved so that \emph{frag\_src} and \emph{frag\_tgt} remain alignable. SMILES are wrapped with the dedicated \texttt{<start\_smiles>}/\texttt{<end\_smiles>} markers introduced in \cref{sec:qwenatom} so that the model can locate molecular spans regardless of surrounding natural language.

\subsection{Downstream task data}

For ChemCoTBench (6 sub-tasks) and QED-DRD2, we use the public training splits as released, converted into the same instruction/input/output schema with SMILES-wrapping normalization and Maximum Common Substructure-based fragment-pair extraction.

\section{Stage~1 internal evaluation: atom-level versus vanilla tokenization}
\label{app:stage1}

We hold the training data, optimization setup, and training steps fixed, and vary only the tokenization scheme. The test set covers 13 held-out Stage~1 sub-tasks, including decomposition, ranking, and vulnerability prediction in both direct and ICL settings.

\begin{table}[!htp]
\caption{Stage~1 internal evaluation: atom-level tokenization versus vanilla tokenization.}
\label{tab:stage1}
\centering
\small
\setlength{\tabcolsep}{4pt}
\resizebox{\columnwidth}{!}{
\begin{tabular}{llccc}
\toprule
Sub-task & Metric & van. & atom & $\Delta$ \\
\midrule
Decomposition           & F1      & 0.880 & \textbf{0.906} & +0.026 \\
Direct Ranking          & K-$\tau$ & 0.760 & \textbf{0.777} & +0.017 \\
Direct Vulnerability    & EM      & 0.798 & \textbf{0.870} & +0.072 \\
ICL Ranking             & K-$\tau$ & 0.626 & \textbf{0.753} & +0.127 \\
ICL Vulnerability       & EM      & 0.600 & \textbf{0.722} & +0.122 \\
MMP Direct Rank         & K-$\tau$ & 0.183 & \textbf{0.216} & +0.033 \\
MMP Direct Vuln.        & EM      & 0.272 & \textbf{0.332} & +0.060 \\
MMP ICL nd. Rank        & K-$\tau$ & 0.453 & \textbf{0.512} & +0.059 \\
MMP ICL nd. Vuln.       & EM      & 0.506 & \textbf{0.540} & +0.034 \\
MMP ICL d. Rank         & K-$\tau$ & 0.472 & \textbf{0.503} & +0.031 \\
MMP ICL d. Vuln.        & EM      & 0.454 & \textbf{0.556} & +0.102 \\
Direct Rank top-1       & Acc     & 0.752 & \textbf{0.770} & +0.018 \\
Direct Rank top-3       & Acc     & 0.892 & \textbf{0.907} & +0.015 \\
\bottomrule
\end{tabular}
}
\end{table}

\end{document}